\documentclass[letterpaper, 10pt, conference]{ieeeconf}
\IEEEoverridecommandlockouts
\usepackage{epsfig}
\usepackage{amssymb}
\usepackage{cite}
\usepackage{booktabs}
\usepackage{multirow}
\usepackage{subcaption}
\usepackage{mathtools}
\usepackage{siunitx}
\usepackage{comment}
\usepackage{threeparttable}
\usepackage{bm}
\usepackage{url}
\usepackage{color}
\usepackage {makecell}

\newcommand{\figlab}[1]{\label{fig:#1}}
\newcommand{\figref}[1]{Fig.~\ref{fig:#1}} 
\newcommand{\tablab}[1]{\label{tab:#1}}
\newcommand{\tabref}[1]{Table~\ref{tab:#1}} 
\newcommand{\forlab}[1]{\label{for:#1}}

\newcommand{\etal}{\textit{et~al.}}

\usepackage{amssymb}
\usepackage{pifont}
\newcommand{\cmark}{\ding{51}}%

\usepackage{amsmath}
\usepackage{algorithmicx}
\usepackage[ruled]{algorithm}
\usepackage[noend]{algpseudocode}

\usepackage{array}

\begin{document}
\title{Affordance-Guided Dual-Armed Disassembly Teleoperation for Mating Parts}
\author{Gen Sako$^{1}$, Takuya Kiyokawa$^{*1}$, Kensuke Harada$^{1,2}$, Tomoki Ishikura$^{3}$, Naoya Miyaji$^{3}$, and Genichiro Matsuda$^{3}$
\thanks{$^{1}$Department of Systems Innovation, Graduate School of Engineering Science, Osaka University, Toyonaka, Osaka, Japan. *Corresponding author: kiyokawa.takuya.es@osaka-u.ac.jp}%
\thanks{$^{2}$Industrial Cyber-physical Systems Research Center, The National Institute of Advanced Industrial Science and Technology (AIST), 2-3-26 Aomi, Koto-ku, Tokyo, Japan.}%
\thanks{$^{3}$Manufacturing Innovation Division, Panasonic Holdings Corporation, 2-7 Matsuba-cho, Kadoma, Osaka, Japan.}%
}

\maketitle

\begin{abstract}
Robotic non-destructive disassembly of mating parts remains challenging due to the need for flexible manipulation and the limited visibility of internal structures. This study presents an affordance-guided teleoperation system that enables intuitive human demonstrations for dual-arm fix-and-disassemble tasks for mating parts. The system visualizes feasible grasp poses and disassembly directions in a virtual environment, both derived from the object's geometry, to address occlusions and structural complexity. To prevent excessive position tracking under load when following the affordance, we integrate a hybrid controller that combines position and impedance control into the teleoperated disassembly arm. Real-world experiments validate the effectiveness of the proposed system, showing improved task success rates and reduced object pose deviation.
\end{abstract}

\IEEEpeerreviewmaketitle

\section{Introduction}
With the increasing severity of environmental issues and resource depletion, the realization of a sustainable society has become imperative.
Remanufacturing, which promotes the reuse of products, has garnered significant attention as a key strategy for sustainable production and consumption~\cite{Sundin2012}.
In particular, while the demand for remanufacturing large home appliances is growing, the disassembly process remains heavily dependent on manual labor.
Labor shortages and safety concerns have driven efforts to automate disassembly operations using robotic systems~\cite{Poschmann2020}.

Mating parts, which are widely used in appliance structures, are typically designed for easy assembly but pose significant challenges for non-destructive disassembly due to the limited visibility of their internal structures~\cite{Yoshida2020}.
In robotic disassembly tasks, determining feasible grasp poses and disassembly directions is critical for reliable execution~\cite{Kiyokawa2024}.
However, due to the lack of an established teleoperation framework for teaching, recent advances in imitation learning techniques have been difficult to apply effectively in this context~\cite{Zare2024}.
Furthermore, variations in product designs and changes in part conditions during the disassembly process add to the complexity of automation.

To address these challenges, this study proposes an affordance-guided teleoperation system that enables effective human demonstration of dual-arm disassembly operations.
During the fixation and disassembly tasks, the system estimates feasible grasp poses and disassembly directions based on the geometric model of the target object.
These affordances are visualized in a virtual environment to support the demonstrator’s decision-making, even when internal structures are occluded and complex.
\figref{overview} shows the overview of our proposed system.

In disassembly tasks, the visualized affordance of disassembly direction can sometimes lead to demonstration trajectories that result in excessive position tracking and force application.
To mitigate this issue, the proposed system incorporates a hybrid control scheme that combines position tracking with impedance control, preventing excessive loads on the object during disassembly.

We validated the system through real-world dual-arm teleoperation experiments.
Its effectiveness was assessed by comparing it with comparison methods in terms of task success rate and object pose deviation during and after disassembly operations.
\begin{figure}[tb]
    \centering
    \small
    \includegraphics[width=\linewidth]{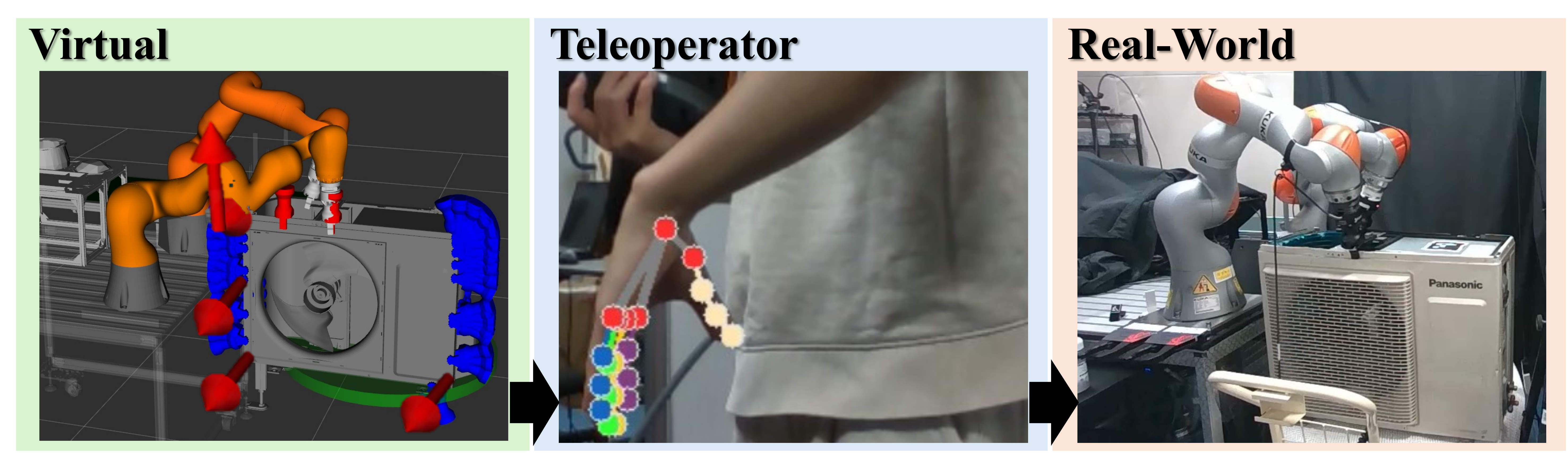}
    \caption{\small{Overview of Proposed Methods}}
    \figlab{overview}
\end{figure}

\section{Related Works}
\subsection{Force-Aware Contact-rich Manipulation}
Several studies have addressed contact-rich manipulation tasks such as assembly and disassembly.
Since disassembly involves frequent contact interactions, force control plays an important role.

Schumacher~\etal~\cite{Schumacher2013} proposed a method for identifying and removing mating parts in small products using force sensing, but their approach is not applicable to large home appliances.
Peternel~\etal~\cite{Peternel2015} implemented intuitive teaching using tactile feedback with position control, but this still led to assembly-target object damage.
Zhang~\etal~\cite{Zhang2019} analyzed the end-effector compliance required for peg disassembly, and Goli~\etal~\cite{Goli2024} achieved peg disassembly using reinforcement learning.

Nevertheless, directly applying these methods to large home appliances remains challenging due to the increased number and variety of mating parts. The difficulties mainly stem from severe occlusions and complex structural configurations inherent to such products.

Humans adapt end-effector impedance to achieve appropriate contact~\cite{Burdet2001}.
The integrated approach of leader-follower teaching, bilateral control, and imitation learning has recently attracted renewed attention~\cite{Talasaz2017,Sakaino2022,Zhao2023,Minelli2023,Rastegarpanah2024}.
While small leader-follower arms are commonly used, larger arms are necessary for handling large home appliances due to payload constraints.
Psomopoulou~\etal~\cite{Psomopoulou2020} utilized a small 6-axis leader and a large 7-axis follower, but they identified challenges related to training requirements and operational usability.

Instead of relying on such complex systems for the user interface, this study proposes a direct teaching method that recognizes demonstrators' movements through image-based recognition.
More importantly, making operators focus on visualized affordance to ease the disassembly teleoperation, even if there are occlusions and structural complexity.

\subsection{Affordance-Based Manipulations}
Many researchers have explored the use of affordances in teleoperation to reduce the cognitive load associated with robotic manipulation tasks. In particular, affordance templates (ATs)~\cite{Hart2015} and affordance primitives (APs)~\cite{Pettinger2022} have been employed to support safe and efficient execution of complex contact tasks across various robotic platforms, including humanoid robots~\cite{Hart2015,Pohl2020,Penco2024} and mobile manipulators~\cite{Pettinger2022,Frank2023}. These frameworks reduce operator burden by abstracting task definitions and autonomously handling low-level control, such as force constraints and motion planning.

ATs define object-centric, user-specified manipulation goals through graphical fixtures, enabling semi-autonomous execution. APs build upon this concept by applying screw theory to further abstract task specifications, such as screwing or pushing directions, resulting in more robust and computationally efficient planning. The AP framework also improves tolerance to misalignments, enabling reliable operation in unstructured environments.

Gorjup~\etal~\cite{Gorjup2019} demonstrated a semi-autonomous teleoperation system for grasping and disassembly, which uses ATs and proposes affordance-based guidance to improve success rates in cluttered scenarios with a humanoid robot.

In contrast to these studies, this study focuses on identifying and leveraging effective affordances specifically suited for dual-arm disassembly of large, complex-shaped objects, aiming to achieve successful and safe teleoperated manipulation.

\section{Proposed Methods}
\subsection{Overview}
\begin{figure}[tb]
    \small
    \centering
    \begin{minipage}[tb]{\linewidth}
        \centering
        \includegraphics[keepaspectratio, width=\linewidth]{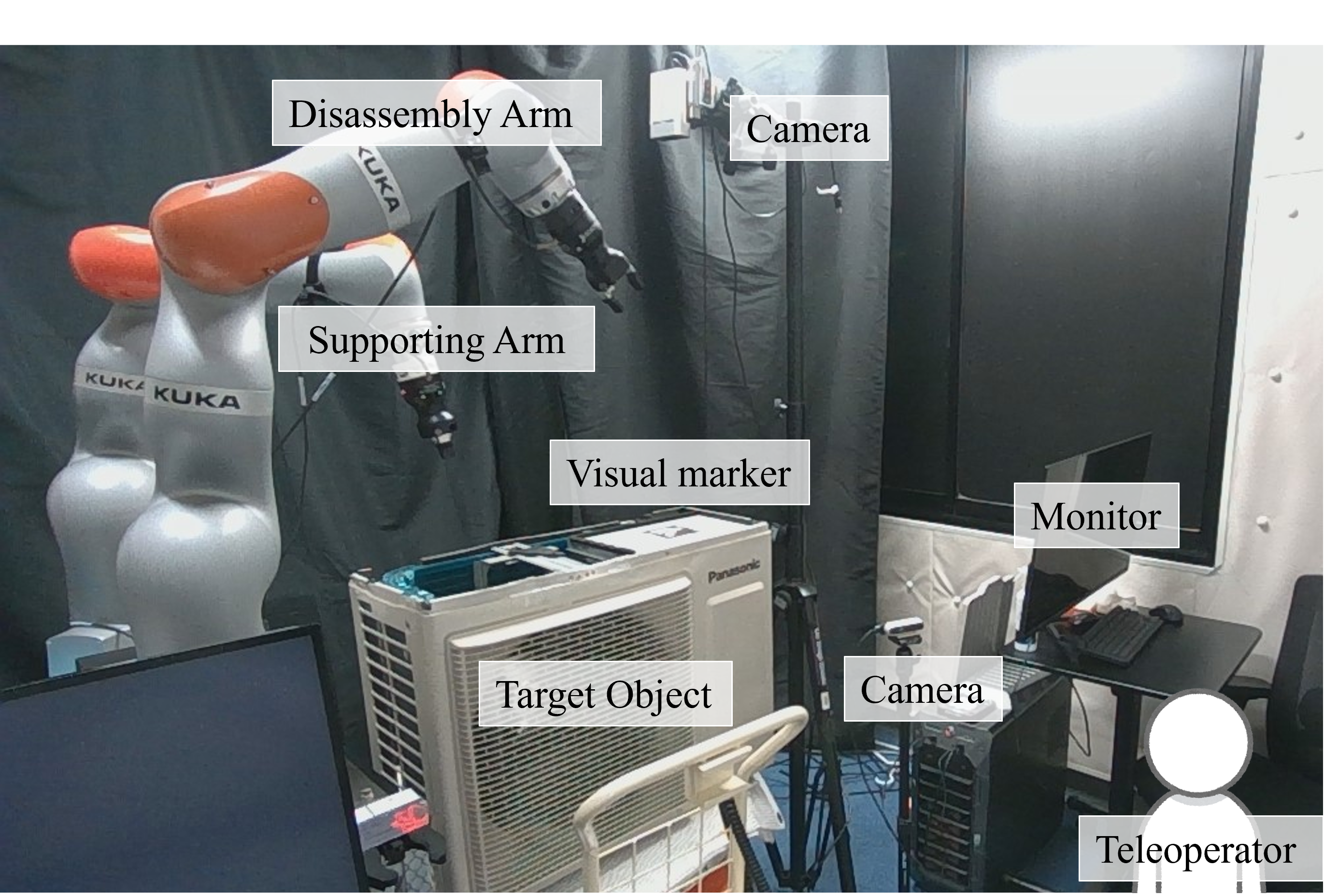}
        \subcaption{\small{Environment}}
    \end{minipage}
    \begin{minipage}[tb]{\linewidth}
        \centering
        \includegraphics[keepaspectratio, width=\linewidth]{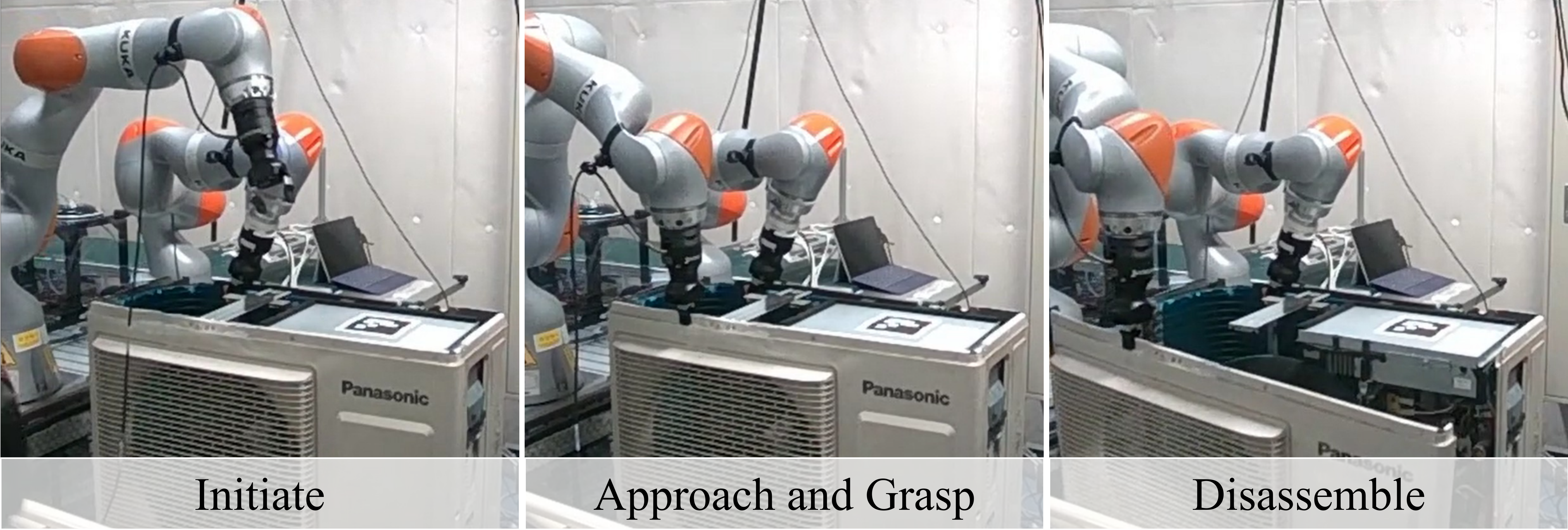}
        \subcaption{\small{Operation}}
    \end{minipage}
    \vspace{0.5mm}
    \caption{\small{Assumed Robotic Disassembly Scenario}}
    \figlab{assumption}
\end{figure}
\figref{assumption}~(a) shows the assumed environment.
A dual-arm manipulator equipped with two two-finger grippers is used to perform both fixation and disassembly tasks on the target object.
In this study, one arm is dedicated to fixation, while the other is responsible for performing the disassembly task.
The demonstrator observes both the physical and virtual environments, and hand movements are captured using an RGB-D camera.

\figref{assumption}~(b) shows the operation flow.
First, the demonstrator approaches the target object using the fixation arm and selects an appropriate grasp pose based on the visualized affordance.
When the gripper enters a user-defined vicinity around a candidate grasp pose, it automatically aligns itself to the corresponding grasp configuration.
Once the object is grasped, the fixation arm is locked to stably hold the object throughout the disassembly process.

Next, the demonstrator uses the disassembly arm to select a grasp pose suitable for the disassembly operation, again guided by the affordance visualization.
The disassembly direction is estimated from the geometry of the target part.
To avoid applying excessive force or damaging the object, the disassembly motion is executed under impedance control, allowing the robot to imitate the demonstrator's motion in a compliant manner.

\subsection{Teaching by Demonstration}
\begin{figure}[tb]
    \small
    \centering
    \includegraphics[width=\linewidth]{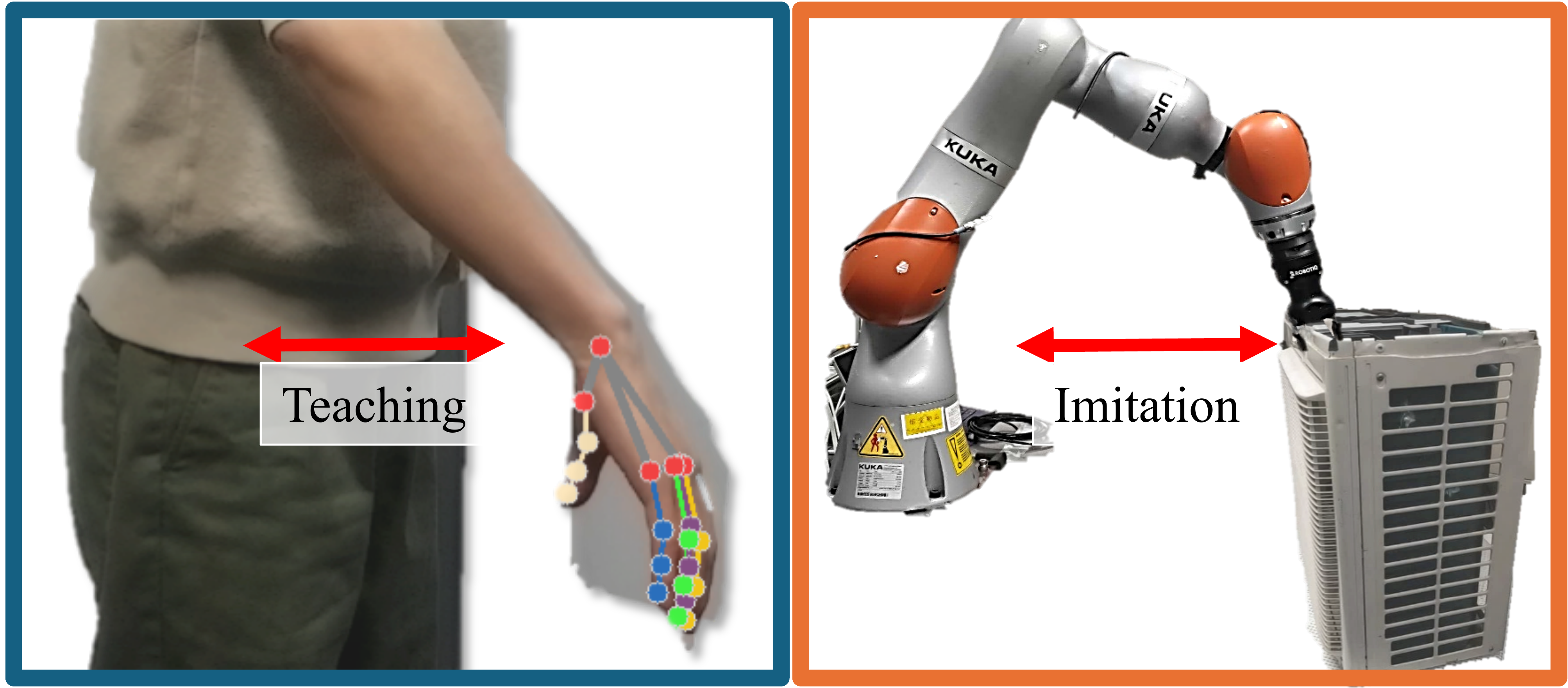}
    \caption{\small{Teaching by Demonstration}}
    \figlab{imitation}
\end{figure}
\figref{imitation} illustrates a demonstration of a disassembly trajectory.
The proposed teleoperation framework, based on teaching by demonstration, is used for both the fixation and disassembly arms.

The demonstrator’s hand motion is captured via human pose estimation to enable demonstration-based teleoperation.
The input RGB-D image is resized and normalized, and a lightweight neural network is applied to detect the hand region and its key points, which correspond primarily to joint locations.
Among the detected key points, only those that remain consistently visible to the camera are selected for use in the teaching process.
The depth values associated with these points allow estimation of the demonstrator’s three-dimensional hand pose.

To define the hand pose, the wrist is designated as the origin.
The $x$-axis is defined as the bisector of the angle formed by the vectors connecting the base joints of the index finger and thumb, directed from the thumb to the index finger, as shown in \figref{imitation}(a).
The positive $y$-axis is aligned with the camera’s depth direction, and the $z$-axis is computed as the cross product of the $x$- and $y$-axes.
The resulting hand pose in the camera coordinate system is then transformed into the corresponding end-effector pose for the robotic arm via a calibrated transformation matrix.

\subsection{Affordance for Feasible and Efficient Disassembly}
\subsubsection{Feasible Grasps}
\begin{figure}[tb]
    \small
    \centering
    \includegraphics[width=\linewidth]{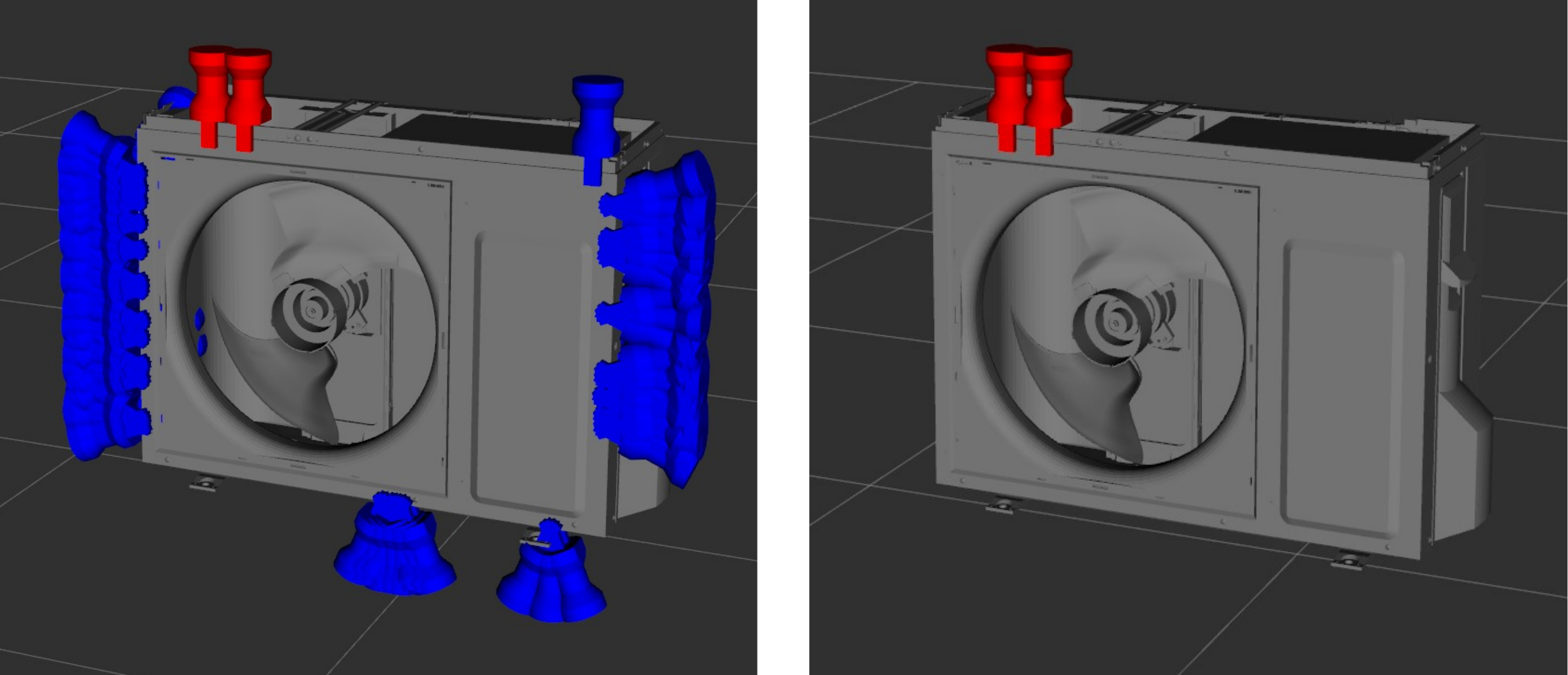}
    \caption{Stable Grasps Based on Geometric Constraints}
    \figlab{grasps}
\end{figure}
We adopt the method proposed by Wan~\etal~\cite{Wan2020} to estimate stable grasp candidates based on the geometry of the target object. Given the mesh models of the target component and the gripper, this method computes feasible grasp poses.

The algorithm begins by clustering mesh facets with similar normal directions and sampling contact points across the surface. To eliminate unstable grasps, contact points near facet boundaries or those positioned too closely are discarded. For a two-finger gripper, the method computes a corresponding contact point in the opposite direction of each surface normal to form a contact pair. The grasp center is defined at the midpoint of the pair, and the algorithm determines a gripper orientation that ensures collision-free contact.

The left side of~\figref{grasps} shows initial grasp candidates. The right side shows how candidates that probably collide with non-disassembly targets are excluded from the final selection.

To choose an appropriate grasp for each task objective, the system evaluates a kind of similarity between each candidate and the estimated hand pose of the demonstrator. The system removes grasp candidates whose positional and orientational differences are both above predefined thresholds.
The positional difference is calculated using Euclidean distance, and the orientation difference is derived from the quaternion inner product, linearly mapped such that ${0^{\circ}}$ corresponds to 0 and ${180^{\circ}}$ to 1. When both criteria are satisfied, the robot autonomously executes the grasp.

\subsubsection{Disassembly Direction}
\begin{figure}[tb]
    \centering
    \small
    \begin{minipage}[tb]{0.65\linewidth}
        \centering
        \includegraphics[keepaspectratio, width=\linewidth]{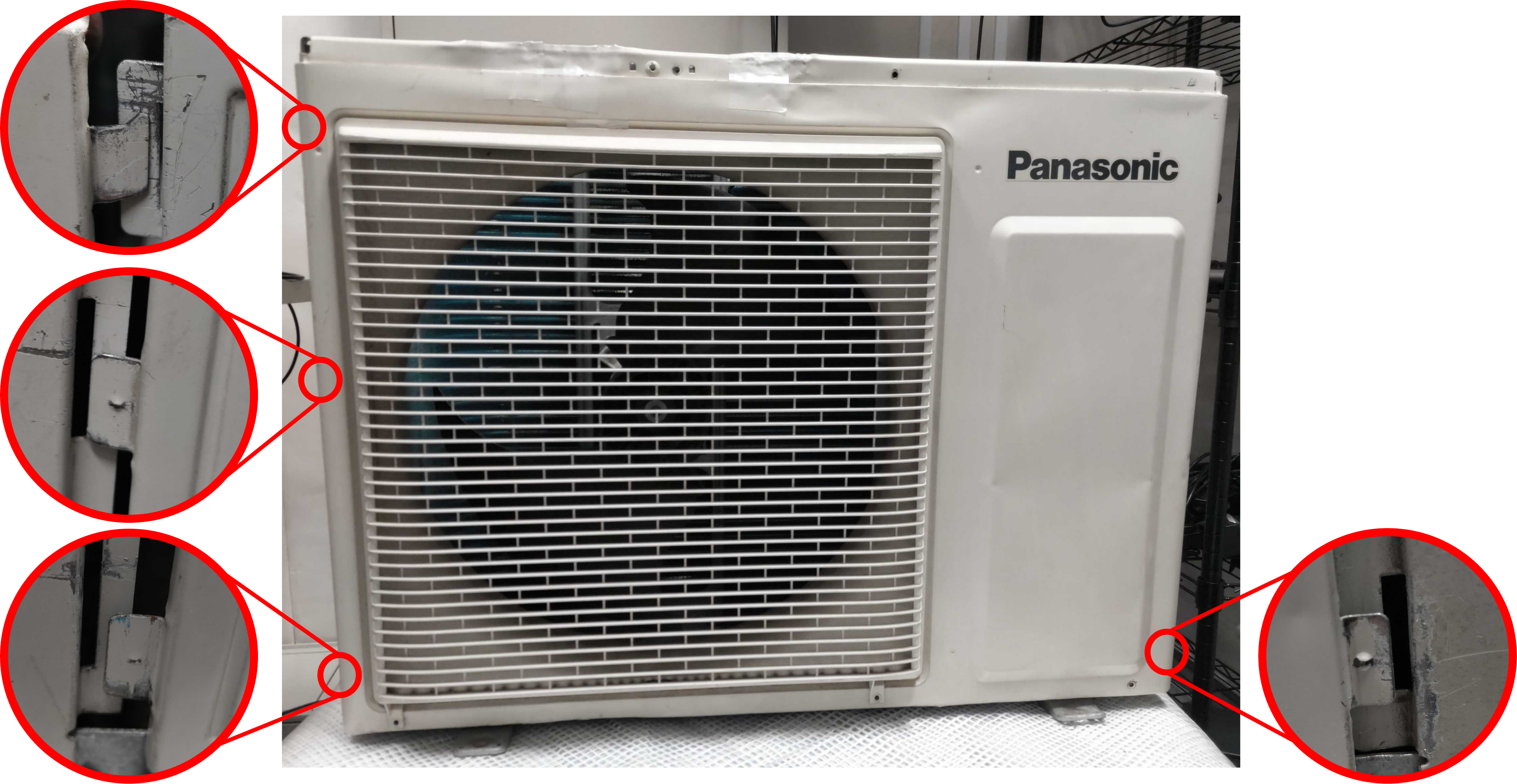}
        \subcaption{\small{Actual Parts}}
    \end{minipage}
    \begin{minipage}[tb]{0.33\linewidth}
        \centering
        \includegraphics[keepaspectratio, width=\linewidth]{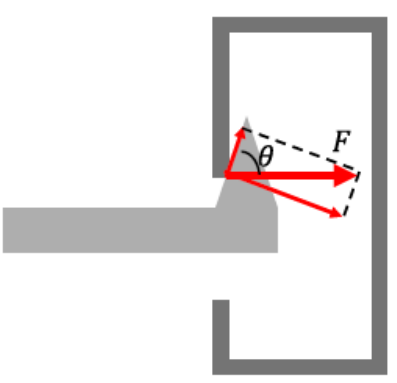}
        \subcaption{\small{Snapfit Hook}}
    \end{minipage}
    \vspace{0.5mm}
    \caption{\small{Mating Parts Existing on Front Cover and Snapfit Hook}}
    \figlab{mating_parts}
\end{figure}
For the disassembly arm, guiding the demonstrator to generate trajectories that apply force in the appropriate direction can reduce cognitive load by helping them more efficiently identify a feasible and effective path.
Since mating parts exhibit some flexibility, assuming fixed-end conditions leads to unrealistic results. To address this, we adopt the method proposed by Suri~\etal~\cite{Suri2000}, which models in-plane stiffness, out-of-plane stiffness, and rotational stiffness as spring elements.

\figref{mating_parts}~(a) shows the pictures of example mating parts.
\figref{mating_parts}~(b) illustrates the assumed dynamics model of each connection.
${\boldsymbol{\theta}}$ indicates the angle between the direction of the projection of the mating part and the horizontal direction. A horizontal force $F$ is applied to the mating part, and its horizontal and vertical parts are calculated as $F\cos{\boldsymbol{\theta}}$ and $F\sin{\boldsymbol{\theta}}$, respectively, based on the angle ${\boldsymbol{\theta}}$.

As ${\boldsymbol{\theta}}$ approaches $\pi/2$, the required force increases; conversely, as it approaches 0, disassembly becomes easier. This allows for adjustment of both the deformation force and the extraction force. Given the high stiffness of metallic mating parts in large home appliances, a stronger extraction force is typically applied in the estimated disassembly direction.

The left of~\figref{overview} shows an example of the estimated disassembly direction for the front cover of the air conditioning unit, which is visually indicated to the demonstrator in a virtual environment. The horizontal direction is shown using an arrow originating from the mating parts, while the diagonal direction is represented with two arrows, the second of which extends from the tip of the first to improve visibility.

\subsubsection{Imitation Teaching Using Impedance Control}
To ensure both safety and adaptability during disassembly, we introduce an impedance-based correction to conventional position control.
Although the affordance provides efficient disassembly directions, strict position tracking along these directions may result in excessive contact forces and potential damage to the object.

Inspired by human manipulation strategies~\cite{Burdet2001}, we employ impedance control using a mass-spring-damper model.
This model enables compliant behavior by dynamically adjusting the reference position based on external forces.

Let $\hat{x}(t)$ denote the nominal target position and $x(t)$ the displacement computed from the impedance dynamics. The model is given by:
\begin{align}
\forlab{impedance}
    M \ddot{x}(t) + D \dot{x}(t) + Kx(t) &= F(t).
\end{align}
where $M$, $D$, and $K$ are the mass, damping, and stiffness parameters, and $F(t)$ is the external force. The correction term $x(t)$ reflects the compliant deviation under force input.

The adjusted reference position $x_{\mathrm{ref}}(t)$ is then computed as:
\begin{equation}
x_{\mathrm{ref}}(t) = \hat{x}(t) + x(t).
\end{equation}
This results in the following overall control law:
\begin{equation}
\label{eq:full_control_law}
x_{\mathrm{ref}}(t) = \hat{x}(t) + \left( M \ddot{x}(t) + D \dot{x}(t) + K x(t) \right)^{-1} F(t).
\end{equation}
This formulation enables force-aware position tracking, allowing the robot to adaptively apply compliant motion during disassembly tasks.

As shown in~\figref{impedance}, when the end-effector moves from $t_0$ to $t_1$, the target pose is updated accordingly, and the impedance controller applies force $f(t_1)$ and torque $\tau(t_1)$ to the object.

If the demonstrator’s hand then moves to a new pose at $t_2$, while the robot remains near $t_0$, the target is suddenly updated to $t_2$, resulting in larger force $f(t_2)$ and altered torque direction $\tau(t_2)$ due to greater displacement and orientation change.

Pure position control cannot absorb such abrupt changes, often causing excessive forces, lifting, or slippage. In contrast, impedance control adjusts the response based on motion resistance, reducing the risk of failure during disassembly.
\begin{figure}[tb]
    \small
    \centering
    \includegraphics[width=\linewidth]{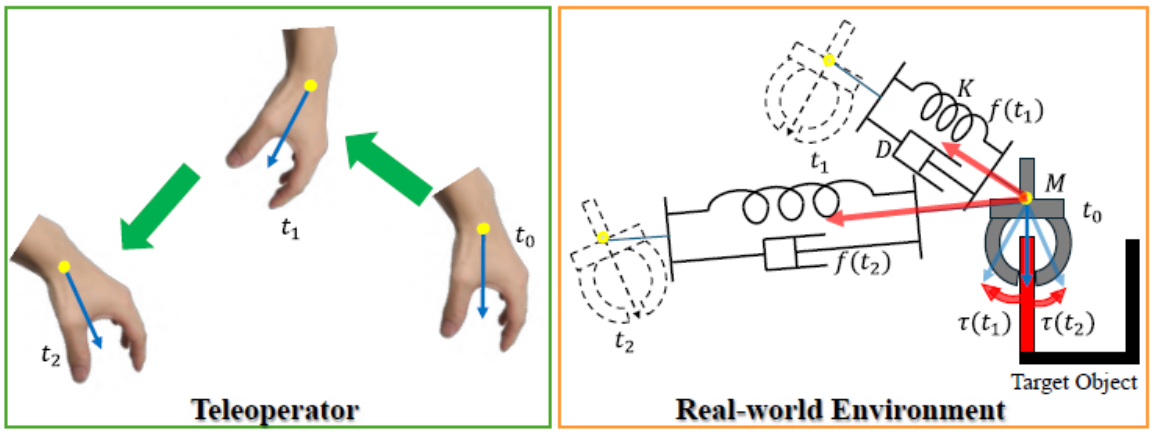}
    \caption{\small{Impedance Control During Disassembly}}
    \figlab{impedance}
\end{figure}

\section{Experiments}
\subsection{Overview}
\begin{figure}[tb]
    \centering
    \small
    \begin{minipage}[tb]{0.49\linewidth}
        \centering
        \includegraphics[keepaspectratio, width=\linewidth]{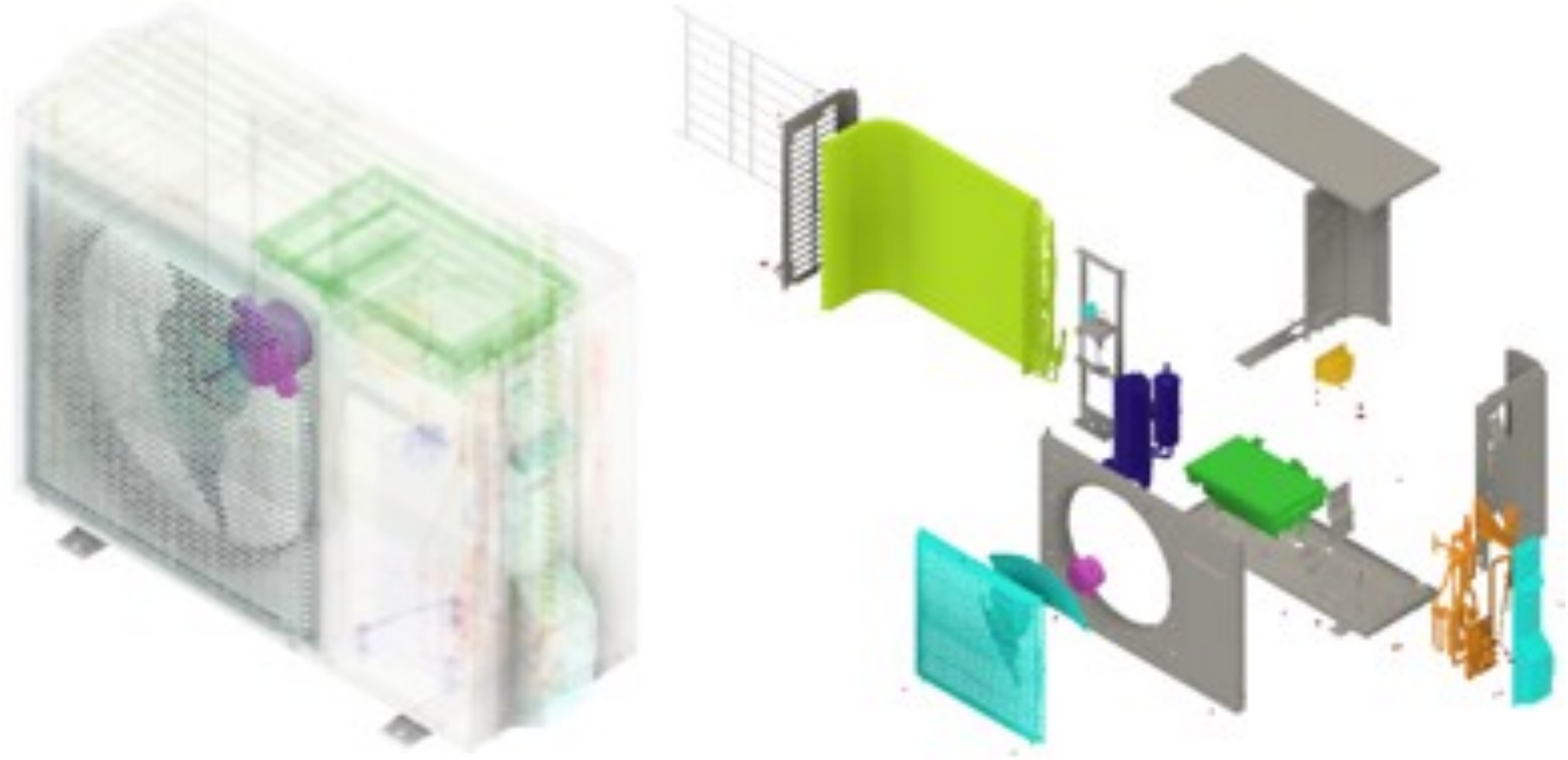}
        \subcaption{\small{All Parts}}
    \end{minipage}
    \begin{minipage}[tb]{0.49\linewidth}
        \centering
        \includegraphics[keepaspectratio, width=\linewidth]{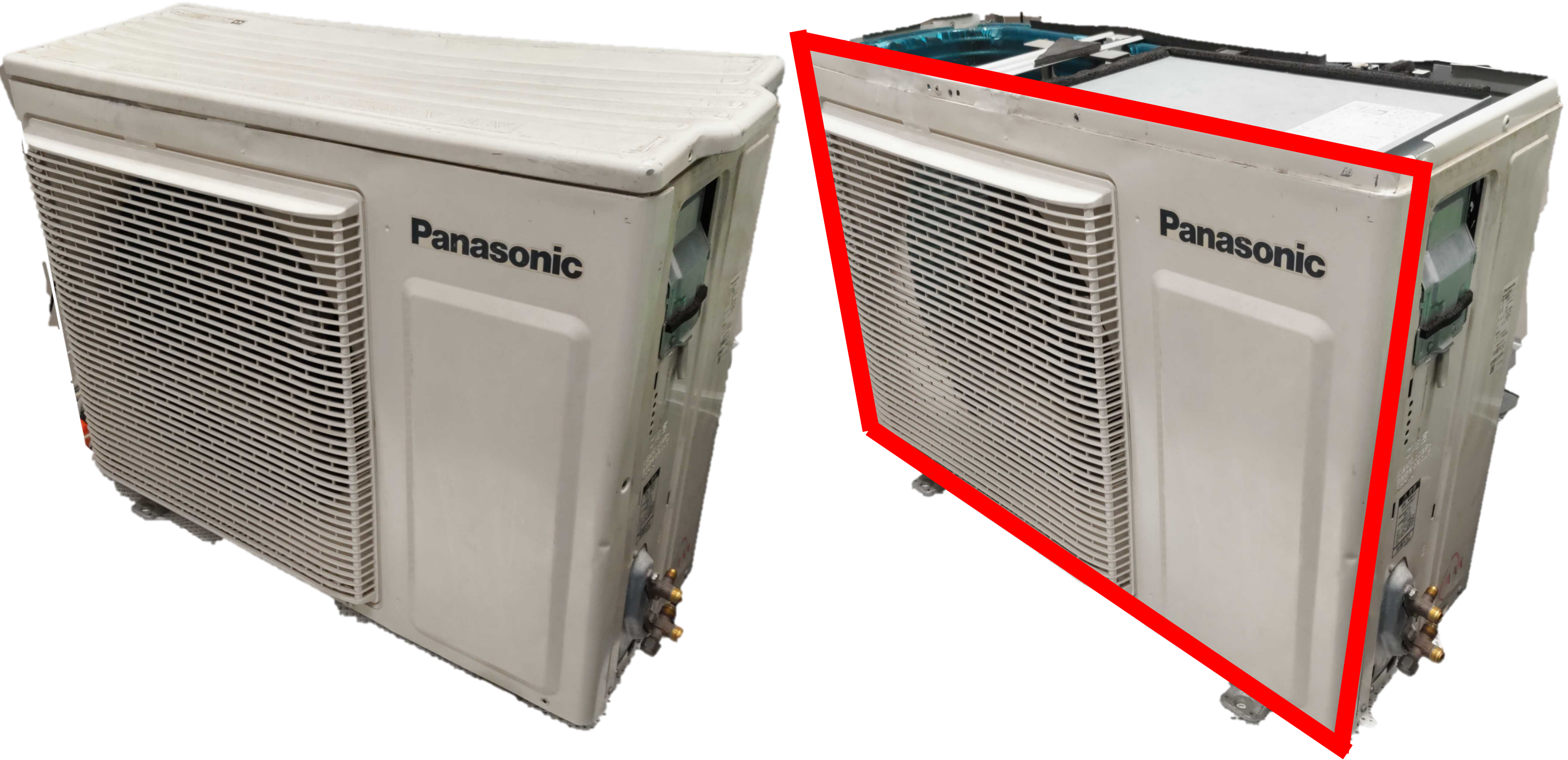}
        \subcaption{\small{Front Cover}}
    \end{minipage}
    \vspace{0.5mm}
    \caption{\small{Target Object (Condenser Unit)}}
    \figlab{object}
\end{figure}
To assess the effectiveness of the proposed method, comprising dual-arm-based fix-and-disassemble operation, affordance-driven teleoperation, and impedance control guided by demonstration, we evaluated the disassembly success rate and object pose deviation.
These metrics were measured under conditions both with and without the use of dual-arm operation, affordance presentation, and impedance control.

\figref{object}~(a) shows the large home appliance target object, which is an air conditioning unit manufactured and used in our experiments.
\figref{object}~(b) shows the target parts; the front cover needed to be disassembled at a recycling factory, requiring non-destructive motor extractions.
In the disassembly, it is hard to remove the front cover, which includes several mating parts of different types.

For the experiment, we used two KUKA LBR iiwa 14 R820 (7-DOF) robotic arms, capable of human-like motions, and two Robotiq Hand-E (two-finger) grippers for their ease of use.
Teaching by demonstration was realized using MediaPipe~\cite{Lugaresi2019}, and hand poses were precisely captured using a WayPonDEV Femto Mega ORBBEC (4K) camera.

\subsection{Evaluation Metrics}
\subsubsection{Task Success Rate}
To evaluate the repeatability, each method was tested in 10 trials, and each trial was determined as a success or failure.

Failures were defined as any of the following: the object slipping from the gripper, breakage of the gripper mount, or exceeding the arm’s range of motion.

\subsubsection{Object Pose Deviation}
To evaluate the fixation performance of the dual arm, we observed the object pose deviation during and after the disassembly operations by each method.
Although the disassembly durations varied, all videos were temporally scaled to 16 seconds to align the timing.

The object pose deviation is calculated using the estimated poses of visual markers as:
\begin{equation}\forlab{deviation}
    ||\Delta\mathbf{p}_t|| + \theta_t,
\end{equation}
where the position deviation $||\Delta\mathbf{p}_t||$ and orientation deviation $\theta_t$ are:
\begin{align}\forlab{deviations}
    ||\Delta\mathbf{p}_t|| &= \sqrt{(x_t-x_0)^2 + (y_t-y_0))^2 + (z_t-z_0))^2}, \\
    \theta_t &= 2 \cos^{-1} (\mathbf{q}_t \cdot \mathbf{q}_0),
\end{align}
where the rotation angle between the quaternion $\mathbf{q}_t$ (the observation at $t$) and the initial pose $\mathbf{q}_0$ represents $\theta_t$.

\subsection{Comparison Methods}
The proposed method incorporates a dual-armed fix-and-disassemble operation, an affordance, and a hybrid controller.

\tabref{comparison} shows the comparison methods we prepared.
Our experiments evaluate three different systems to see the effect of the two different factors: the affordance-guided demonstration with our hybrid controller and dual-armed fix-and-disassembly operations.
\begin{table}[tb]
    \small
    \centering
    \caption{\small{Comparison Methods}}
    \begin{threeparttable}
        \begin{tabular}{lccc} \toprule
            Method & Dual arms & Affordance & Hybrid controller \\ \midrule
            Baseline &  & \cmark & \\
            Comparison &  & \cmark & \cmark \\
            Proposed & \cmark & \cmark & \cmark \\ \bottomrule
        \end{tabular}
        \vspace{0.5mm}
    \end{threeparttable}
    \tablab{comparison}
\end{table}

\subsection{Results}
\subsubsection{Task Success Rate}
The overall operation task success rates for baseline, comparison, and proposed methods are 80\%, 100\%, and 100\%, respectively.
The baseline method had the lowest success rate, followed by the comparison and proposed methods with 100\% success rates.

In the baseline method, grasping failures were more frequent compared to the other two methods, with the most common failure being the dropping of the target object. This can be attributed to the lack of grasping affordance suggestions, which prevented the selection of appropriate grasping positions and orientations. 

\subsubsection{Object Pose Deviation}
\figref{deviation_plot} and \figref{deviation} show the calculated values from the detected markers and their images used.
The deviations in the methods without dual arms are higher than the proposed method over time.
On the other hand, regarding the disassembly time, the trials without impedance control were slightly faster than those with impedance control, but the difference was only 1 or 2 seconds. In all cases, the disassembly was completed within 12 seconds.

This result suggests that dual-armed fix-and-disassembly is possible to reduce the deviation of the target object during the demonstration.
\begin{figure}[tb]
    \small
    \centering
    \includegraphics[width=\linewidth]{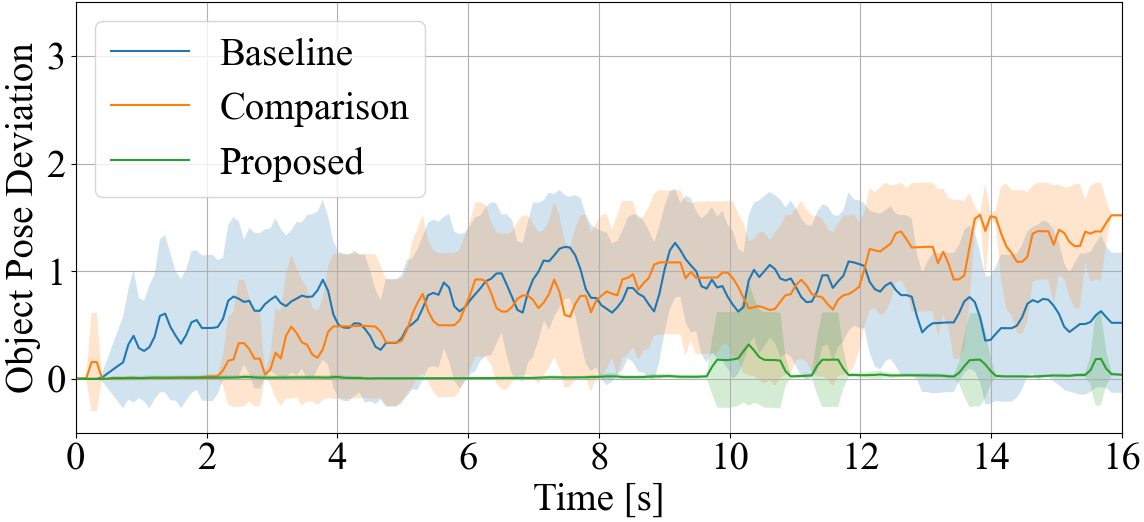}
    \caption{\small{Object Pose Deviation}}
    \figlab{deviation_plot}
\end{figure}
\begin{figure}[tb]
    \small
    \centering
    \begin{minipage}[tb]{\linewidth}
        \centering
        \includegraphics[keepaspectratio, width=\linewidth]{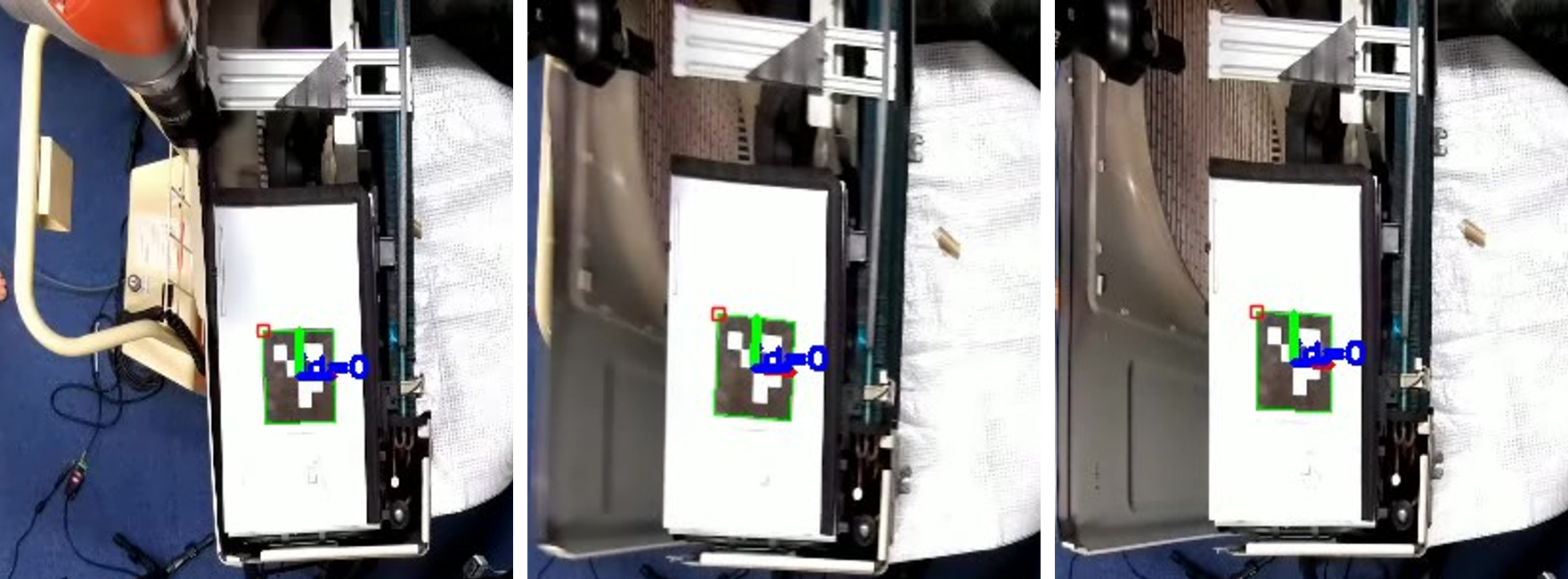}
        \subcaption{\small{Baseline}}
        \vspace{0.5mm}
    \end{minipage}
    \begin{minipage}[tb]{\linewidth}
        \centering
        \includegraphics[keepaspectratio, width=\linewidth]{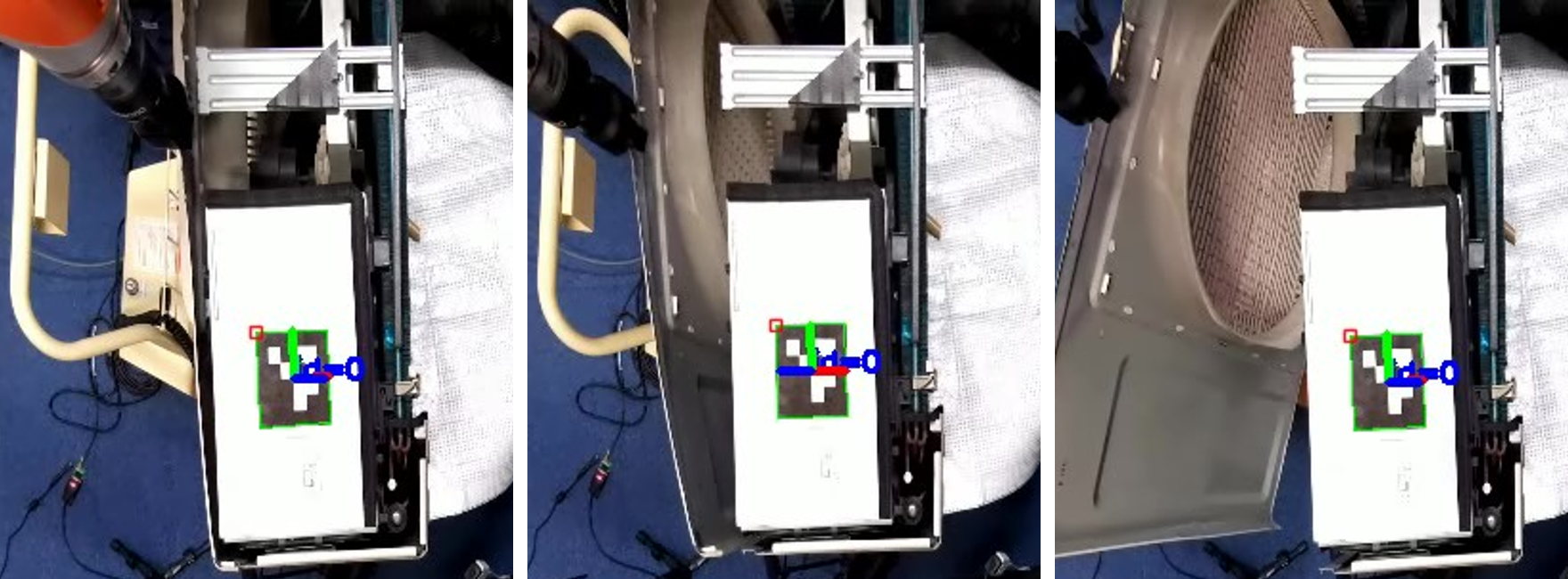}
        \subcaption{\small{Comparison}}
        \vspace{0.5mm}
    \end{minipage}
    \begin{minipage}[tb]{\linewidth}
        \centering
        \includegraphics[keepaspectratio, width=\linewidth]{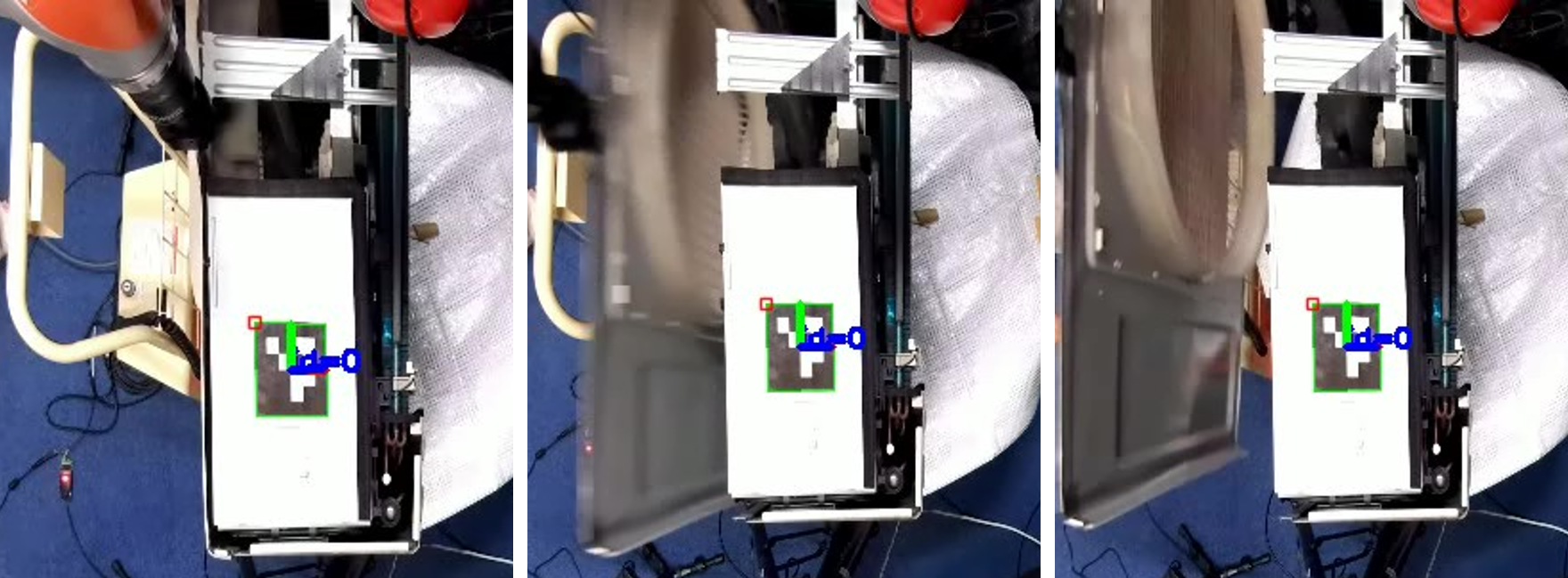}
        \subcaption{\small{Proposed}}
    \end{minipage}
    \caption{\small{Deviations of Objects during Disassembly Operations}}
    \figlab{deviation}
\end{figure}

\section{Discussion}
When comparing failure cases between the proposed method and the baseline, the proposed method prevented gripper and arm mount damage as well as grasping slippage.
This suggests that impedance control effectively reduces excessive force applied to components, facilitating non-destructive disassembly.
These results demonstrate its effectiveness of the hybrid controller, suggesting the potential to suppress excessive position tracking under overload conditions.

Furthermore, benefiting from the fixation of the target object, the other results suggest that dual-armed fix-and-disassembly is possible to reduce the deviation of the target object during the operations.

\section{Conclusion}
To facilitate efficient and non-destructive disassembly of mating parts, such as those found in large home appliances, we proposed an affordance-guided teleoperation framework using dual-arm coordination.
By leveraging object geometry to guide grasp and disassembly directions, and incorporating impedance control into the disassembly arm, the system enables intuitive and compliant manipulation.

Experimental results demonstrated that the proposed method outperformed baseline approaches in terms of disassembly success rate and reduced object pose deviation. The hybrid controller effectively mitigated failures caused by excessive force, and the dual-arm strategy enhanced stability during fix-and-disassemble tasks.

Future work includes the automatic tuning of impedance parameters and further validation with a broader range of objects in real-world scenarios.

\section*{Acknoledgement}
This work was supported by the New Energy and Industrial Technology Development Organization (NEDO) project JPNP23002. 

\bibliographystyle{IEEEtran}
\footnotesize
\bibliography{reference}

\end{document}